\pdfoutput=1

\documentclass[11pt]{article}

\usepackage{acl}

\usepackage{times}
\usepackage{latexsym}
\usepackage{graphicx}
\usepackage{textcomp}
\usepackage{bm}
\usepackage{soul,color}

\usepackage[T1]{fontenc}

\usepackage[utf8]{inputenc}

\usepackage{microtype}

\title{GEST: the Graph of Events in Space and Time as a \\Common Representation between Vision and Language}

\author{Mihai Masala \\ mihaimasala@gmail.com
\And  Nicolae Cudlenco \\ cudlenconick@gmail.com 
      \AND Traian Rebedea\\ trebedea@gmail.com  \And Marius Leordeanu \\ leordeanu@gmail.com 
 }

\begin{document}
\maketitle
\begin{abstract}
One of the essential human skills is the ability to seamlessly build an inner representation of the world. By exploiting this representation, humans are capable of easily finding consensus between visual, auditory and linguistic perspectives. In this work, we set out to understand and emulate this ability through an explicit representation for both vision and language - Graphs of Events in Space and Time (GEST). GEST alows us to measure the similarity between texts and videos in a semantic and fully explainable way, through graph matching. It also allows us to generate text and videos from a common representation that provides a well understood content. In this work we show that the graph matching similarity metrics based on GEST outperform classical text generation metrics and can also boost the performance of state of art, heavily trained metrics. 

\end{abstract}

\section{Introduction}
Making connections between vision and language seems easy for humans, but extremely challenging for machines, despite a large body of research on image and video captioning ~\cite{you2016image,aneja2018convolutional,anderson2018bottom,gao2017video,zhou2018end,wang2018reconstruction}, visual question answering~\cite{antol2015vqa,lu2016hierarchical,zhong2020self}, image synthesis~\cite{reed2016generative,dong2017semantic,zhou2019text} or video generation~\cite{li2018video, balaji2019conditional, wu2022nuwa, singer2022make, villegas2022phenaki}. While major improvements were made using Transformers~\cite{vaswani2017attention}, there is still a long way to go.
Also, these tasks were widely tackled independently of each other, with no significant push for a more unified approach. 

For tasks involving vision or language, information is usually processed by an encoder (e.g. Transformers, CNNs or LSTMs) that builds a numerical representation. While this approach is ubiquitous across both vision and NLP, it is fundamentally limited by its implicit, mostly unexplainable, and highly volatile nature. We strongly believe that such a representation can be replaced (or augmented) by a better, explicit, and more robust one.

\begin{figure*}
    \centering
    \includegraphics[width=\textwidth]{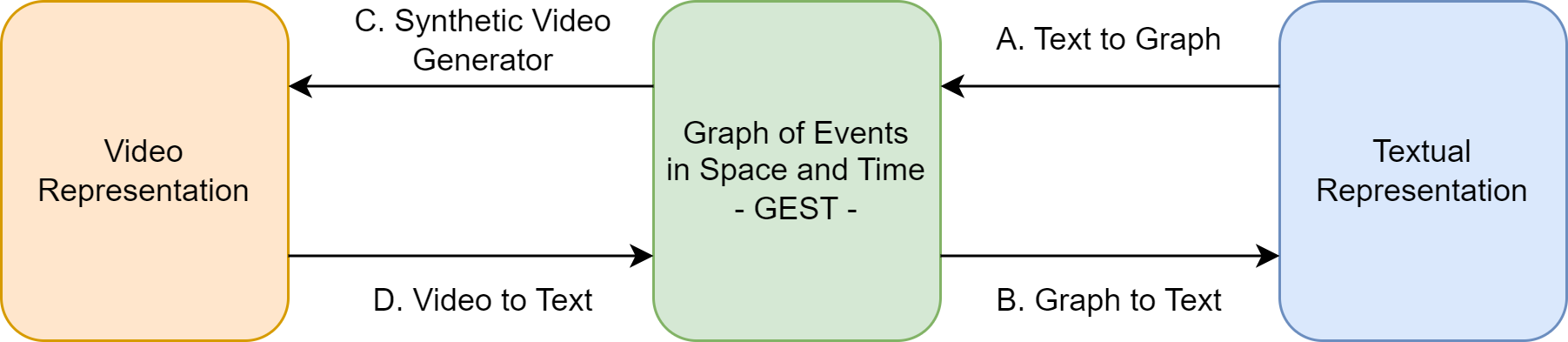}
    \caption{Functional overview of the proposed framework, centered around GEST. GEST represent the central component, allowing for seamless transitions between different forms. For example the transition from text to video is done via steps A and C, while the transformation from video to text can be done via steps D and B. In this work we focus on modules A and B
    .}
    \label{fig:general}
\end{figure*}

In this work we introduce the Graph of Events in Space and Time (GEST) for representing visual or textual stories, as groups of events related in space and time at any level of semantics. GEST provides a common and meaningful representation, through which we can compute similarities or differences between texts or videos, and we could also generate texts or videos in an explainable, analytical way. 


\section{Related Work}

\noindent \textbf{Graphs that model text:}
Graphs were traditionally used in natural language processing (NLP) in many forms:
syntactic trees (e.g. dependency or constituency parsing trees)~\cite{lin1998dependency,culotta2004dependency},
semantic trees (in the form of Combinatory Categorial Grammar)
~\cite{zettlemoyer2012learning}, Rhetorical Structure Theory (RST)~\cite{mann1988rhetorical} trees, Discourse Graphs~\cite{christensen2013towards}, knowledge graphs~\cite{hao2017end,bauer2018commonsense,wang2019explainable} and Abstract Meaning Representation (AMR) graphs~\cite{banarescu2013abstract}. Recently, Graph Neural Networks (GNNs)~\cite{zhou2020graph,wu2020comprehensive} were employed to parse and encode such structures. 
RST trees~\cite{mann1988rhetorical} and Discourse Graphs~\cite{christensen2013towards} were developed as theories of text organization using relations between claims as the central component and emphasizing relations between these claims. Then, knowledge graphs are used for encoding true fact about the world, allowing for efficient interogation for Question Answering systems. Conversely, AMR graphs are semantic and represent links between concepts from the natural text. Crucially, two syntactically different sentences can share the same AMR graph if they are semantically similar.

\textbf{Graphs that model videos:}
Graphs were also used as a way to model videos~\cite{sridhar2010relational, aoun2011graph, singh2017graph}. While previous approaches~\cite{brendel2011learning, chen2016efficient,yuan2017temporal,wang2018videos, cherian20222} consider the nodes in the graph as video regions, we opt for a fundamentally different approach, modeling events as graph nodes. 

\citet{aditya2018image}
define Scene Description
Graphs (SDGs), graph-based intermediate representation built specifically for describing images. SDGs are based on objects, actions and semantic (based on KM-Ontology
~\cite{clark2004km}
), ontological and spatial relations. With GEST we explicitly add the temporal aspect as we are interested in representing videos instead of images. Furthermore, our formulation is uniform (everything is an event), leads to a more compact representation, allows for more complex (e.g. semantic, logical) relations between events, while also being capable of representing such events at different scales (see Figure
~\ref{fig:multiple_scales}).


\textbf{Text generation metrics:}
Text generation metrics were studied in the field of NLP 
for comparing two or more texts or documents~\cite{sai2022survey}. Common metrics include BLEU~\cite{papineni-etal-2002-bleu}, METEOR~\cite{banerjee-lavie-2005-meteor}, ROUGE~\cite{lin-2004-rouge} and 
SPICE
~\cite{anderson2016spice}. While BLEU and ROUGE compute the similarity as the exact n-gram overlap, METEOR uses a more relaxed matching criteria, based on stemmed matching, followed by synonymy and paraphrase matching.
SPICE builds a semantic scene-graph that is used to extract information about objects, attributes and their relationships.
More recently, BERT~\cite{devlin2019bert} was integrated into text metrics.
BERTScore~\cite{zhang2019bertscore} uses a BERT backbone to obtain embeddings for each token in the hypothesis and reference, which are matched using a greedy approach,
The state-of-the-art BLEURT~\cite{sellam-etal-2020-bleurt} is pre-trained on a large number of synthetic samples then finetuned on WMT~\cite{bojar-etal-2017-results} and WebNLG~\cite{gardent-etal-2017-webnlg}. Synthetic data is generated by altering Wikipedia sentences via back-translation, random words dropping or mask-filling with BERT. Most pretraining signals are employed in the form of BLEU, ROUGE and BERTScore, back-translation likelihood or textual entailment.

All mentioned text generation metrics employ clear rules, but they lack explainability, due to the space in which computations are formed. The n-gram space of BLEU, METEOR or ROUGE is simple, but totally counter-intuitive for humans. In the case of BERTScore and BLEURT the projected space is even more blurry and void of any intuitive understanding. Instead of projecting texts into an n-gram or Transformer space, we propose a new representation space, namely the space of events in space-time. Comparing events and their relations expressed in two texts is much more natural. The fact that the GEST space is explicit and grounded in the real world, is the very reason for which we obtain explainability and interpretability.

\section{Graph of Events in Space and Time}
\label{section:GEST}

Fundamentally, a GEST is a means of representing stories. We focus on modeling stories as they are the main way of expressing ideas, sentiments, facts, perceptions, real-world or fantasy happenings. Stories are an essential component in theater, cinema in the form of storyboards and are also an integral part in relating, communicating and teaching historical events. Stories are universal: a life is a story, a dream is a story, a single event is a story. Atomic events create intricate stories in the same way that small parts form an object in a picture, or how words form a sentence. Therefore, in modeling stories, we distinguish interactions in space and time as the central component. In general, changes in space and time lead to the notion of events and interactions. Similarly to how changes in an image (image gradients) might represent edges, space-time changes (at different levels of abstraction) represent events.
Accordingly, events in space and time could be detectable, repeatable and discriminative. Interactions between events in space and time change the current state of the world, can trigger or cause other events and in turn cause other changes. Therefore, we use these events and their interactions in space and time as the fundamental component of GEST. 
Fundamentally, an edge connects two events in space and time. This connection can be, but is not limited to temporal (e.g. after, meanwhile), logical (e.g. and, or) or spatial (e.g. on top of). Since a node in GEST can also represent physical objects (e.g, “The house exists for this period of time”) the graph connections can represent any potential relation between two objects or two events: the event “house” was involved in event: “holding a meeting at that house”. Therefore, an edge can also represent an event by itself
.
For each event we encode mainly the type of action, the involved entities, the location and the timeframe in which an event takes place. Crucially, in GEST both explicit (e.g. actions) and implicit (e.g. existence) events are represented using the same rules. A GEST example can be found in Fig.~\ref{fig:multiple_scales} 
, while more examples are in Appendix, Sec.~\ref{section:further_examples}.

GEST can represent events at different scales, in greater detail by expanding an event node into another graph, or in a lesser detail by abstracting a graph into a single event node. In Fig.~\ref{fig:multiple_scales} we exemplify the power of such an approach. On the left of Fig.~\ref{fig:multiple_scales} we show the GEST associated to the following story: "John says that Daniel bought a watch". In the right half we expand the event "Daniel bought a watch" to a more detailed story (GEST). All other event nodes can be expanded into their own GEST stories (e.g. the paying action can be further expanded by detailing the procedure: currency, amount, method and so on). In principle, any GEST could become an event into a higher-level GEST and vice-versa, any event could be expanded into a more detailed GEST.

\begin{figure}
    \centering
    \includegraphics[width=\columnwidth]{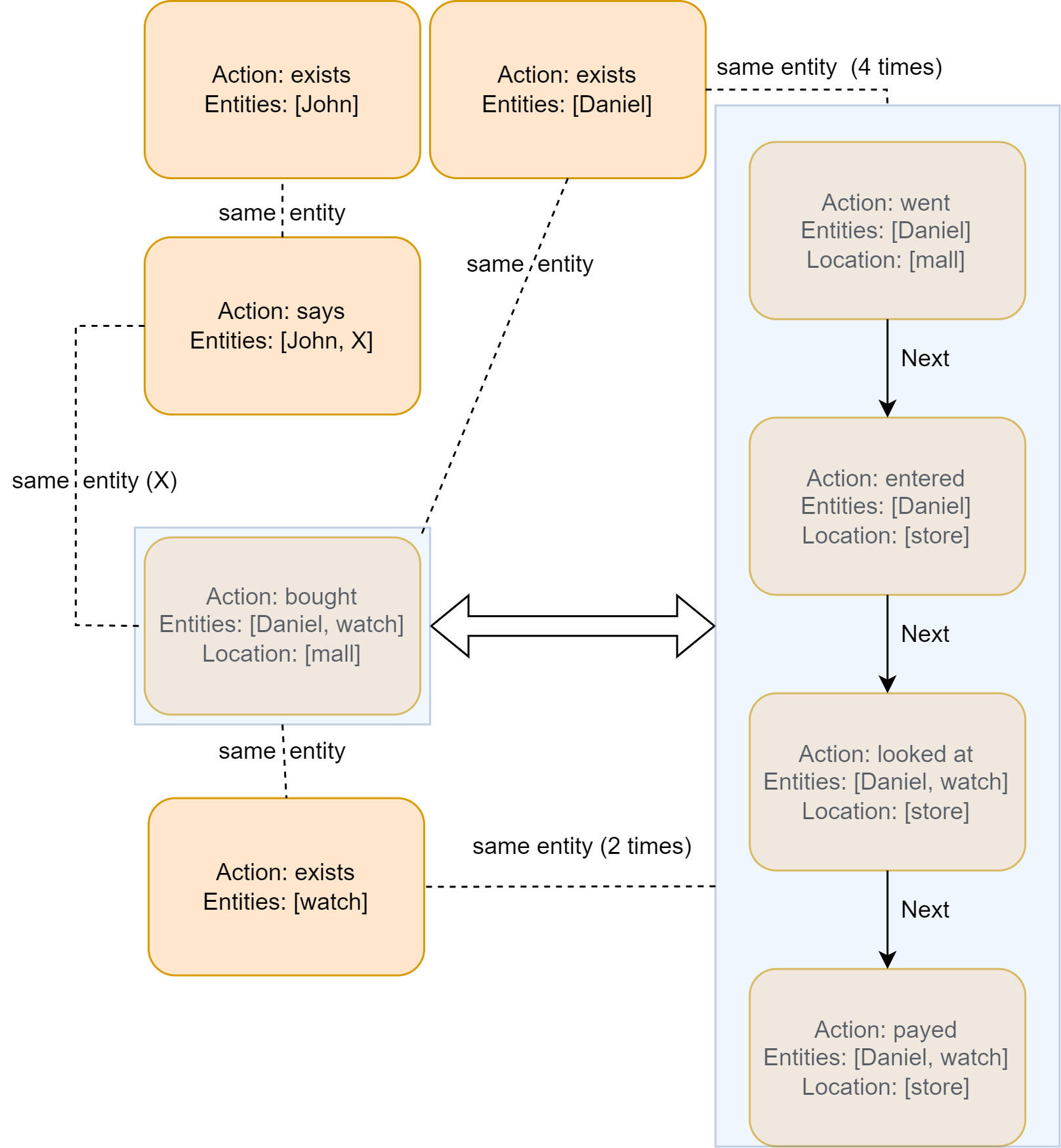}
    \caption{GEST that illustrates the concept of multiple viewpoints and graph-node equivalence. Note that for brevity, we omit some details in the nodes (e.g. timeframe) and also add details to emphasize some points (e.g. the same entity edges).}
    \label{fig:multiple_scales}
\end{figure}

GEST represents concisely what happens in the real world. So, when vision and language represent the same world, they could also be represented by the same GEST. GEST is suitable for many tasks, including video-to-text or text-to-video generation. GEST is an alternative to the standard way of solving these tasks. Instead of generating natural language descriptions directly from an obfuscated and implicit representation given by a video encoder, GEST breaks video captioning in two problems: generate GEST from video, followed by generating text from GEST. Conversely, generating a video starting from a text prompt can be split into building GEST from text, followed by independently creating the video (Fig.~\ref{fig:general}). 
In this paper we demonstrate both directions and the advantages of the approach. We also argue that the main advantage of the highly-explicit GEST representation is to give total knowledge and control over the content of the text or video.  
Additional details and formal definition of GEST is given in Appendix, Sec.~\ref{section:definition}.

\newpage

\textbf{Building ground truth GEST from text:}
Ground truth GEST from text is needed for training and evaluation. We note that building GEST representation from text is not a trivial task, and we aim to automate this process. Nevertheless, to obtain correct GEST from text human intervention is still needed. From each sentence, we want to extract information such as the type of actions, the entities involved, locations and the times of actions, as well as their relations.
All this is extracted by parsing the dependency tree (automatically extracted\footnote{https://spacy.io/models/en\#en\_core\_web\_lg last accessed on 19th of January 2023}) of each individual sentence using a set of handcrafted rules (followed if needed by human correction). Context (e.g. location inference) and event ordering is also injected into the graph to obtain the complete GEST of a story.

\subsection{bAbI corpus}

The bAbI corpus~\cite{weston2015towards} introduces a set of 20 question answering tasks that are designed to act as proxy tasks for reading comprehension. As the grammar of bAbI is rather simple, we devised a set of handcrafted rules to automatically parse the dependency tree of each sentence in order to extract the relevant information. For bAbI, the text-to-graph automatic module works flawlessly, always detecting and extracting the correct information from each sentence.
In this work we focus on bAbI tasks numbered 1, 2, 3, 5, 6, 7, 8, 9, 10, 11, 12, 13, 14. We leave the other tasks for future work, as they are devised with other goals in mind (e.g. tasks numbered 16 and 18 are devised for basic induction or size reasoning) 
This leads to a total of 26.364 graphs, with 21.588 train, 2.388 validation and 2.388 test graphs.

\subsection{Videos-to-Paragraphs dataset}

The Videos-to-Paragraphs dataset~\cite{bogolin2020hierarchical} contains videos with two stages of text representations. The 1st stage contains
contains simple sentences that describe simple actions, while the 2nd stage contains semantically rich descriptions. This duality is especially suited for GEST as the 1st stage is simple enough that we can immediately extract events as simple actions. The 2nd stage is semantically richer and compacter. This represents a crucial step-up from the bAbI corpus where only the simpler linguistic stage is present. Following~\cite{bogolin2020hierarchical}, we will refer to sentences from the 1st stage as SVOs (Subject, Verb, Object) and to the 2nd stage texts as stories. In Videos-to-Paragraphs we identify thre types of temporal relations between events (SVOs): "next", "same time" and "meanwhile", using soft margins to extract them. Using both 1st and 2nd stage texts annotations, we build (with minimal manual intervention) ground truth GEST representations for the entire dataset, a total of 1048 samples (with a 85-5-10 training, dev, test split) consisting of GESTs and the two stages of text descriptions. 

\section{GEST as a metric for comparing stories}

We first want to study and evaluate the power of GEST to capture content from stories in natural language. Ideally, different texts that illustrate the same underlying story should have the same GEST. We evaluate this property by first defining a similarity metric between two GESTs and compare its performance (in separating texts that represent the same story vs. different stories) to other metrics from the literature that work directly on the original text in natural language.  

\begin{table}
\centering
\begin{tabular}{lcccc}
\hline

\textbf{Method} & \textbf{Corr} &  \textbf{Acc} & \textbf{F} & \textbf{AUC} \\

\hline

BLEU@4 & 24.45 & 75.52 & 0.2816 & 52.65\\
METEOR & 58.48 & 84.23 & 1.1209 & 73.90\\
ROUGE & 51.11  & 83.40 & 0.7164 & 68.92\\
SPICE & 59.42 & 84.65 & 1.0374 & 74.43 \\
BS & 57.39 & 85.89 & 1.0660 & \underline{77.93}\\
\hline
G SM& \underline{61.70} & 84.65 & \underline{1.2009} & 75.47 \\
G NGM & 60.93 & \underline{86.31} & 0.9770 & 76.75 \\

\hline
BLEURT & \textbf{70.93} & \textbf{90.04} & \textbf{2.0280} & \textbf{88.02} \\
\hline
\end{tabular}
\caption{Results comparing GEST representation power with common text generation metrics applied on stories from Videos-to-Paragraphs test set. Both text generation metrics and graph similarity function are applied on the ground truth (stories and graphs). We show in \textbf{bold} the best value for each metric, and with \underline{underline} 2nd best. BS stands for BERTScore, G for GEST, corr for correlation, Acc for Accuracy, F for Fisher score and AUC for the area under the precision-recall curve. For brevity all (except F) are scaled by 100.}
\label{table:metrics_story_vs_classic}
\end{table}

\subsection{Graph matching similarity metric}

Comparing two GEST representations, being graphs, is naturally suited for a graph matching formulation. we test two graph matching methods, a classical approach, Spectral Matching (SM)~\cite{leordeanu2005spectral} and a modern deep learning based approach, Neural Graph Matching (NGM)~\cite{wang2021neural}. SM is a fast, robust and accurate method that uses the principal eigenvector of an affinity matrix\footnote{more details on building the matrix in Appendix, Sec. \ref{section:experimental_setup_metric}}, while NGM employs multiple neural networks that learn and transform the affinity matrix into the association graph, which is further embedded and used as input for a vertex classifier. 

\subsection{Results and Discussion}

Results in Tab.~\ref{table:metrics_story_vs_classic} attest the power of our proposed representation: graph matching in the GEST space outperforms all classic text generation metrics (i.e. BLEU@4, METEOR and ROUGE) and even modern metrics based on pre-trained Transformers such as BERTScore. Nevertheless, the specifically and heavily trained BLEURT metric outperforms all considered metrics on this dataset. Note that the other metrics all lack access to the sheer amount of data that BLEURT metric was trained on (around 1.8 million samples). We reckon that given such data, a trained GRAPH-BLEURT metric could outperform the original BLEURT.

\begin{table}
\centering
\begin{tabular}{lcccc}
\hline
\textbf{Method} & \textbf{Corr} &  \textbf{Acc} & \textbf{F} & \textbf{AUC} \\
\hline
BLEURT & 70.93 & 90.04 & 2.0280 & 88.02 \\
\hline
+BLEU@4 & 70.93 & 90.04 & 2.0274 & 88.04\\
+METEOR& 71.20 & 89.63 & 2.0659 & 87.62\\
+ROUGE & 70.76 & 90.04 & 1.9973 & 87.71\\
+SPICE & \underline{71.94} & 88.80 & \underline{2.0808} & 87.71\\
+BS& 71.11 & 89.63 & 2.0089 & 87.25\\
\hline

+G SM& \textbf{72.89} & \textbf{90.87} & \textbf{2.2086} & \textbf{89.80}\\
+G NGM& 71.91 & \underline{90.46} & 2.0537 & \underline{88.58}\\


\hline
\end{tabular}
\caption{Results comparing the power of BLEURT coupled with common text generation metrics and GEST, applied on stories from Videos-to-Paragraphs test set. Text generation metrics are computed on the ground truth stories, while the GEST similarity (G) with graph matching is computed on GEST learned from stories. Notations are the same as in Tab. ~\ref{table:metrics_story_vs_classic}.}
\label{table:metrics_story_learned}
\end{table}

The initial test show the representational power of GEST, but they do not test yet the capability of this representation to be combined with a heavily trained one. That would be another, complementary way, to prove the effectiveness of GEST. We test this capability by showing that GEST can boost a state-of-the-art, strongly trained metric, even when we combine the two in the simplest, linear way. Starting from the original text of the story, we learn to transform the story automatically into GEST, and then obtain a GEST similarity score between stories by comparing, using graph matching, the corresponding generated GESTs. A second, BLEURT score between the stories is obtained as before. We then learn, on the training set, how to linearly combine the two scores, to best separate the texts of the same story vs. texts of different stories.
We apply the same procedure to all classic metrics, in order to evaluate the benefit brought by GEST relative to other methods.
We learn to transform a graph from a story in natural text, by using a  sequence-to-sequence framework, with the story as input and the serialized graph as output. For further details on the training process see Appendix, Sec.~\ref{section:experimental_setup_metric}. 

In Tab.~\ref{table:metrics_story_learned} we show the results of BLEURT (top), those of other metrics combined with BLEURT using the same linear regression approach (middle) and the results of GEST (bottom), using the two graph matching methods (SM and NGM). It is important to note that in combination with other metrics BLEURT does not always improve, but when combined with GEST it always improves and by the largest margin. In the Appendix
Sec.~\ref{section:graph_vs_bleurt}, we show cases when BLEURT fails to 
predict when two different textual descriptions stem from the same video. In the first case this is due to the different writing style of the two annotators, while in the second case BLEURT assigns a high similarity score in spite of the fact that different actors perform somewhat similar actions. In both cases, the graph matching algorithm manages to correctly predict if the two pairs depict the same video. These tests prove the power of GEST: its new space and associated graph matching metric can be effectively used, with minimal training cost, to boost the performance of existing state-of-the-art.

\section{GEST for text generation}

\begin{table*}
\centering
\begin{tabular}{lccccccccc}
\hline
\textbf{Method} & \textbf{B@1$\uparrow$} & \textbf{B@2$\uparrow$} & \textbf{B@3$\uparrow$} & \textbf{B@4$\uparrow$} & \textbf{M$\uparrow$} & \textbf{R$\uparrow$} & \textbf{C$\uparrow$} & \textbf{BS$\uparrow$} & \textbf{BT$\uparrow$}\\
\hline
S2T  & 43.81 & 30.95 & 22.87 & 16.90 & 20.15 & 38.87 & 78.60 & 38.87 & 58.28\\
G2T & 46.73 & 32.90 & 24.23 & 18.15 & 20.88 & 39.57 & 87.29 & \textbf{41.24} & 58.73\\
\hline
S2T\footnote[2]{}
 & 42.39 & 30.37 & 22.73 & 17.21 & 20.32 & 39.64 & \textbf{96.10} & 40.63 & 57.57\\
G2T\footnote[2]{} & \textbf{52.34} & \textbf{36.92} & \textbf{27.11} & \textbf{19.91} & \textbf{23.18} & \textbf{41.49} & 94.59 & 40.42 & \textbf{59.42}\\
\hline
\end{tabular}
\caption{Results for the task of the text generation on the test set of Videos-to-Paragraphs dataset, presented using common text generation metrics: BLEU@N (B@N), METEOR(M), ROUGE(R), CIDEr(C), BERTScore(BS) and BLEURT(BT). In the S2T (SVOs to text) experiments we trained models that take as input the SVO sequence, while in the G2T (Graph to text) experiments we give the serialized graph as input. \footnote[2]{} marks experiments in which we use an additional training stage with data from bAbI corpus. We highlight with \textbf{bold} the best value for each metric. For brevity all values are scaled by a factor of 100.}
\label{table:text_gen_results}
\end{table*}

GEST describes the world in terms of events and how they relate in space and time and could provide a common ground between the real space and time and "what we say" about it in natural language. Atomic events in a linguistic story (e.g. SVOs) are also well formed events in real space and time, thus they provide a direct link between both worlds. Then relations between events define the space-time structure at semantic level, inevitably becoming a central component in natural language generation. In the following set of experiments we want to better understand and evaluate the importance of these relations, which are an essential component of GEST. We will evaluate the importantce of these connections between events, by comparing language that is generated from events only (task
S2T - SVOs-to-Text)
to language that is generated from events and their relations, that is full GESTs (G2T - GEST-to-Text) - for both using the sequence-to-sequence net. 

We perform the tests on the Video-to-Paragraphs dataset, where the relations between events are mainly temporal in nature. Thus, to better highlight the differences between the textual SVOs and GEST representations
we decide to break the implicit temporal relations given by SVOs ordering, by randomizing (with the same seed) both representations. In the case of SVOs, the order is randomized while for the graphs the order of the edges in the representation is randomized (based on the SVOs permutation). In this setup we can clearly evaluate the impact of the temporal information encoded in the graph structure.

\subsection{Results and Discussion}

Results in Tab.~\ref{table:text_gen_results} validate that GEST is suited for text generation and provides a better representation than plain textual descriptions of the atomic events. Conceptually, the graph representation should always be better as it explicitly encodes complex temporal relations that are not present in SVOs. Nevertheless this does not directly guarantee a better off the shelf performance for text generation as the available training data in our tests is very limited. Our tests show that these
limitation is overcome by the power of the representation. In the first two rows of Tab.~\ref{table:text_gen_results}, both SVOs to text (S2T) and graph to text (G2T) models are trained starting from a general pre-trained encoder-decoder model with no previous knowledge of our proposed representation. Even in this very limited setup (under 900 training samples) the graph representation proves to be superior. Adding more pretraining data, using the bAbI corpus only extends the performance gap between the two approaches (last section of Tab.~\ref{table:text_gen_results}). In the case of bABi we only have access to a single textual representation for each graph, which is akin to the SVOs in the Videos-to-Paragraphs dataset. For this reason, the S2T task on bAbI can be simply solved by using the identity function, while the G2T task can be solved by describing each node. However they provide valuable aditional pretraining data, especially for G2T as it helps the model to better understand and order events in time. The ability to understand and order events in time enables a better transition from simple sentences to longer, more complex natural language

\section{Conclusions}

In this paper we introduce GEST, which could set the groundwork for a novel and universal representation for events and stories. We discuss and motivate its necessity and versatility, while also empirically validating its practical value, in comparing and generating stories. Even with very limited data, our experiments show that GEST is more than fitted for recreating the underlying story, within a space that allows for very reliable and human correlated comparisons. 
This explicit and structured nature of the GEST space lends itself beautifully to various other uses (e.g. video generation).

GEST aims to bring together vision and language, as a central component in an explainable space. Such explicit models are largely missing in the literature, but as we believe that our work demonstrates, they could be useful to better understand language and also control its relation to the visual world.

\section{Limitations}
Maybe the most important limitation of our work is, for the moment, data availability. This lack of quality data affects both learning tasks (i.e. graph-to-text and text-to-graph), as access to more graph-text pairs will greatly improve performance. We found that this is especially relevant for the text-to-graph task, as we conjecture that is represents the harder task. Because we use models that are pre-trained with natural language as input and output, the graph representation has to be learned and understood by the encoder (for the graph-to-text task) and the decoder (for the text-to-graph task). Especially with a limited number of samples, we believe understanding the new graph representation is easier than generating it. Moreover, for the text-to-graph task we ask the decoder to generate a very structured output, defined by a precise grammar.

Our experiments highlighted the power of GEST when applied on real-world events with temporal relations between events. Crucially, this represents only a small subset of what GEST can model. Due to lack of data, we are unable in this work to show the full potential of GEST, namely to represent more abstract events. For example, a revolution is still an event, but at the same time a complex story comprised of multiple events.




\bibliography{custom}
\bibliographystyle{acl_natbib}

\appendix

\section{More Examples of GEST}
\label{section:further_examples}

\begin{figure}
    \centering
    \includegraphics[width=\columnwidth]{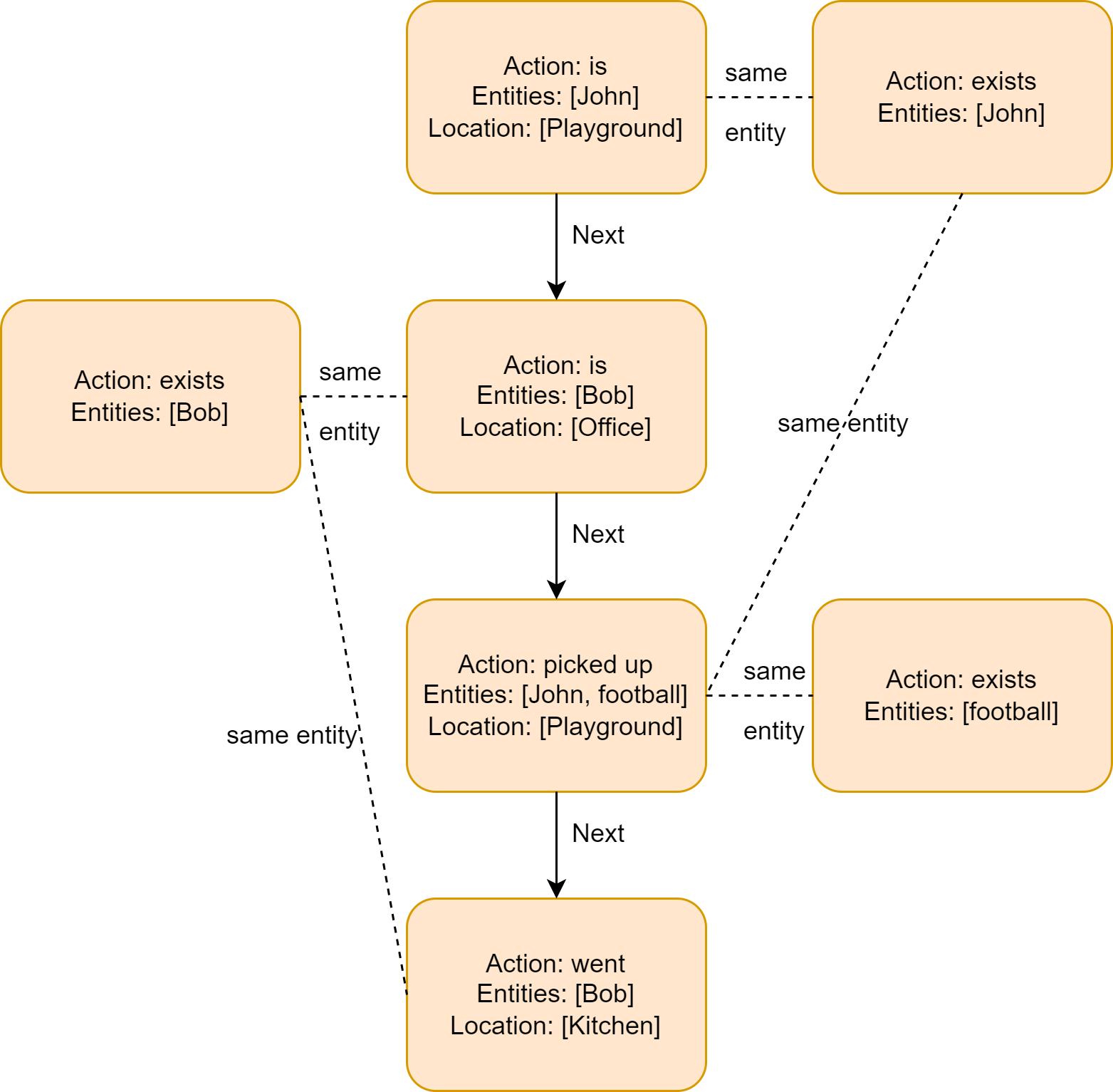}
    \caption{Automatically generated GEST associated to the following bAbI story: "John is in the playground. Bob is in the office. John picked up the football. Bob went to the kitchen.". Notice that the third sentence does not contain any explicit location. Nevertheless, this location is inferred from the context of the story and is added in the corresponding node. Also note that for brevity, we omit some details in the nodes (e.g. timeframe and location for some nodes) and also add details to emphasize some points (e.g. the "same entity" edges).}
    \label{fig:babi_example_multi}
\end{figure}

Figure~\ref{fig:babi_example_multi} presents an example of automatically generated GEST based on a story from the bAbI corpus. Furthermore, this example is used to showcase the spatio-temporal inference (in this particular case only spatial inference) aspect present in converting textual stories into GEST: the node corresponding to the sentence "John picked up the football" contains no explicit spatial information (as the sentence contains no such information), but the location is inferred from the previous action of entity John.

\section{Definition of GEST}
\label{section:definition}
Formally, a Graph of Events in Space and Time (GEST) is a graph defined by the following components:

\bm{$V$} = set of event nodes

\bm{$E$} = set of edges, $\textbf{E} \bm{\subseteq} \textbf{V X V}$

where 

\bm{$V_i$} = ($action$, $entities$, $location$, $timeframe$, $properties$)

\bm{$E_i$} = temporal (we use~\citet{allen1981interval} time interval algebra with minor modifications for ease of use), spatial (e.g. on top, behind, left of), logical (e.g. and, or, cause/effect, double implication) or semantic relations.

Finally, each element of $V_i$ is defined as follows:

\bm{$action$} = the main action; string

\bm{$entities$} = list of entities that are involved in the action; [string]

\bm{$location$} = list of locations in which the action takes place; [string]

\bm{$timeframe$} = list of timeframes in which the action takes place; [string]

\bm{$properties$} = additional properties; dict <property:value>

All elements of \bm{$V_i$}, with the exception of the $properties$ field, can also refer to other nodes. In particular, in the case of $entities$ we use references to the "exists" node for each actor or object involved in an action (this is emphasized by the added dotted lines in the graphical representation). For complex cases, such as the one presented in Figure~\ref{fig:multiple_scales}, references to other actions can be used in $entitites$ to model complex interactions (including multiple viewpoints). 

Furthermore, our GEST framework is a step-up from the classic Subject-Verb-Object (SVO) approach. In our case, the Subject becomes an event (even if we are talking about events of type "exists", they are still events) and also the Object becomes an event. An event is composed by objects, and any event requires interaction between objects and the world. As in our formulation objects are events, any interaction (and so any edge) becomes in itself an event. This allows a hierarchical and recursive representation in GEST. Classic models represent object to object interactions, that GEST can easily represent as well. Moreover, we can go to the next level, modeling hyper-events, collapsing such interactions to a single node, generating an infinite recursive process in which nodes expand and collapse into events.

\section{Building ground truth GEST from text}

 In the case of bAbI dataset, inferring the location or timeframe in which an action takes places (for actions that do not explicitly provide this information) is simply done by memorizing the last place or time in which a given entity (mainly actor) was found. This simple process works for this particular dataset as all changes in time or space are explicitly present in text: each movement of any entity in space is always marked with a sentence (e.g. John travelled to the kitchen, Mary went to the office), while if explicit timeframes are mentioned in a sentence, they are mentioned in all sentences of a given story (this happens for task number 14, the only task in which timeframes are mentioned). For task number 14 we apply an additional sorting step before parsing, to ensure that sentences are in chronological order (i.e. we consider the following chronological order "yesterday", "this morning", "this afternoon", "this evening" and break ties using the original order in the story).

While for the bAbI corpus all entities (e.g. actors, objects) are unique for each story (e.g. a single actor with the name John in each story), in the case of Videos-to-Pargraphs this is not always the case. In this case we have to manually intervene and set the proper references (build and link the proper number of nodes), as different entities are referred with the same name in the SVOs (e.g. "man", "desk"). To find and accurately annotate these cases we manually go through each pair of SVOs and story and semantically check their validity. To ease this process we define a set of personal objects (e.g. phone, cup, backpack), entities that are intimately linked with their owner. Unless other specified, all personal objects are unique (in the sense that each owner has its own unique personal object) and, for example, two phones linked with different actors (e.g. by the action of "speaking at the phone") will be represented using two different nodes. We give special attention to cases in objects (personal or not) are passed around different actors. For example, for the following set of SVOs "John picks up his backpack. John gives the backpack to Mary" we will define a single node representing the backpack and internally keep track of its owner. So if a future SVO (in the same story) tells that "Mary handed the backpack to Michael" we will not define a new node and use the reference to the previously defined backpack (as the original backpack moved from John to Mary to Michael). We are now left with cases in which the same word, or set of words, refer different entities. In most cases the telling sign is the presences in the story (not in the SVOs!) of the word "another". For example if the SVOs are "A man talks at the phone. A man enters the room." and the story is "A man talks at the phone. After that, another man enters the room", we will build and use two different nodes (one for each "man') to properly represent the story. For these cases we manually annotate if and where the same word refers to an already existing entity or we have to build a new one. Nevertheless this happens for around 5\% of the entries (i.e. 49 out of 1048).

\section{Experimental Setup}

\subsection{Text Generation Metrics}
For fair comparisons across the paper we employ a board set of text generation metrics: BLEU~\cite{papineni-etal-2002-bleu}, METEOR~\cite{banerjee-lavie-2005-meteor}, ROUGE~\cite{lin-2004-rouge}\footnote{We use the common ROUGE-L variant}, CIDEr~\cite{vedantam2015cider}, SPICE~\cite{anderson2016spice}, BERTScore~\cite{zhang2019bertscore}\footnote{With the recommended "microsoft/deberta-xlarge-mnli" bacbkone} and BLEURT~\cite{sellam-etal-2020-bleurt}\footnote{With the recommended "BLEURT-20" checkpoint}. Computing the metrics was done using coco-captions\footnote{https://github.com/tylin/coco-caption last accessed on 8th of December 2022} for BLEU, METEOR, ROUGE, CIDEr and SPICE, and official code released by the authors for BERTScore\footnote{https://github.com/Tiiiger/bert\_score last accessed on 8th of December 2022} and BLEURT\footnote{https://github.com/google-research/bleurt last accessed on 8th of December 2022}.

\subsection{GEST as a metric}
\label{section:experimental_setup_metric}

For this set of experiments we use the Videos-to-Paragraphs dataset, as it contains multiple annotations (at all levels, graph, SVOs and story) for each video. We consider triplets that stem from the same video as positive examples (they provide different representation for the same underlying story presented in video format), while any other pairs of triplets as negative examples. For our experiments we consider all positive pairs from the test set (a total of 67 in our case) and 174 negative pairs sampled randomly from the same test set.  

\begin{figure*}
    \centering
    \includegraphics[width=\textwidth]{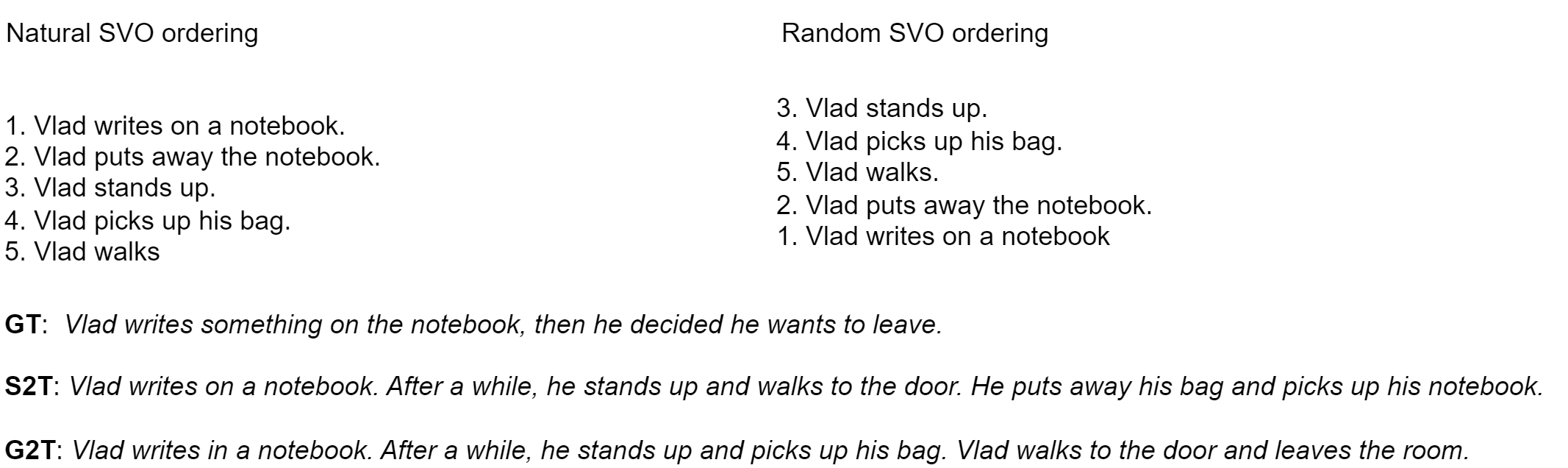}
    \caption{Qualitive example for the story generation task on the test set of Videos-to-Paragraphs dataset. Both natural (as found in the dataset) and randomized SVOs are provided in the left and in the right column. S2T experiments mark models that take as input the SVO sequence (in random order), while for G2T we give the serialized graph as input (in the same random order). Note that, while both trained models output the same number of activities (namely five) the G2T model is capable of better reconstructing the original order, although not perfectly. }
    \label{fig:generation_example_12}
\end{figure*}

For both SM and NGM algorithms, the affinity matrix is built using both node and edge level similarity functions that exploit pre-trained word embeddings. We use pre-trained GloVe~\cite{pennington2014glove} embeddings of size 300, to measure the similarity at each level (e.g. action, entities) for nodes. In order to compare two edges, we integrate node-level similarity (from the nodes that are connected to the particular edges) with the edge-level similarity (i.e. the similarity between the edges type). Essentially, two nodes are as similar as are their actions and entities, while the similarity of two edges is given by multiplying the edge type (e.g. next, meanwhile) similarity with the similarity of the corresponding nodes.

For a fair comparison, the graph similarity metrics is normalized as follows: 

\[\hat{f}(G_1, G_2)=\frac{f(G_1, G_2)}{\sqrt{f(G_1,G_1) * f(G_2,G_2)}}\]

As we are interested in how similar two human annotated stories/graphs are, we need to ensure that all used metrics are symmetric. Out of the considered metrics, only BERTScore and the graph similarity are symmetric by their very own construction. For all other metrics, we force symmetry by computing the metric twice for each pair $(s_1, s_2)$ (i.e. $metric(s_1, s_2)$ and $metric(s_2, s_1)$) and average the results.

For both datasets (i.e. bAbI and Videos-to-Paragraphs) we use the official train, dev and test splits and follow the standard training procedure (i.e. training exclusively on the train set until the performance on the dev split stabilizes and reporting the results on the test set).

For learning to building graphs from raw text, we finetune a GPT3 model (text-curie-001\footnote{https://platform.openai.com/docs/models/gpt-3 last accesed on 8th of May 2023}) on Videos-To-Paragraphs dataset. The model has as input the textual description (story) and the output is represented by the (serialized) graph.
Finally, we apply a syntactic correction step to ensure that the generated strings are well formed and therefore can be converted into a GEST. 


\subsection{GEST for text generation}
\label{section:experimental_setup_to_text}


For text generation we invert the inputs and ouputs from the graph-learning case: the input now is represented by the (serialized) graph, while the output is the ground truth story. As for the model, we start from a pre-trained BART Base~\cite{lewis2019bart} (140M parameters), minimizing the cross-entropy loss using the Adam optimizer~\cite{kingma2014adam}, for a maximum number of 100 epochs. We set the base learning rate at 1e-5 and use a warm-up phase that lasts for 10\% of the entire training process. Training is done on six Nvidia Quadro RTX 5000 graph cards, using an effective batch size of 24 (4 x 6). In this case, for Videos-to-Paragraphs dataset an epoch took around 45 seconds, while 11 minutes per epoch were needed for bAbI. 

\section{GEST for text generation}
\label{sec:appendix_text_gen}



In Figure~\ref{fig:generation_example_12} we present a qualitative example of story generation. This example highlights several important factors. First, as previously mentioned, the story part is not just a description of the atomic actions. The ground truth label for this examples contains only one explicit atomic action present in the SVOs, the writing in a notebook. The other part of the action, namely leaving the room is not explicitly present in the SVOs. Nevertheless preliminary actions, such as packing, standing up, walking are present and with enough examples a model can (and does!) learn that such actions usually entail leaving. However, in this particular case, the models are not able to summarize the actions into a single action (i.e. leaving). The S2T model has some problems reconstructing the original order of actions and also misjudges some entities, inverting the bag and the notebook in the picking up and putting away actions. On the other hand, the G2T model generates a more coherent order of action that better matches the original order. While better, the implicit order generated by the G2T model is still not perfect: it "misses" an action (the second action in the natural order and next to last in the randomized one). Missing to describe an action does not necessarily represent an error as it can represent a part of summarization process.



\begin{figure}
    \centering
    \includegraphics[width=\columnwidth]{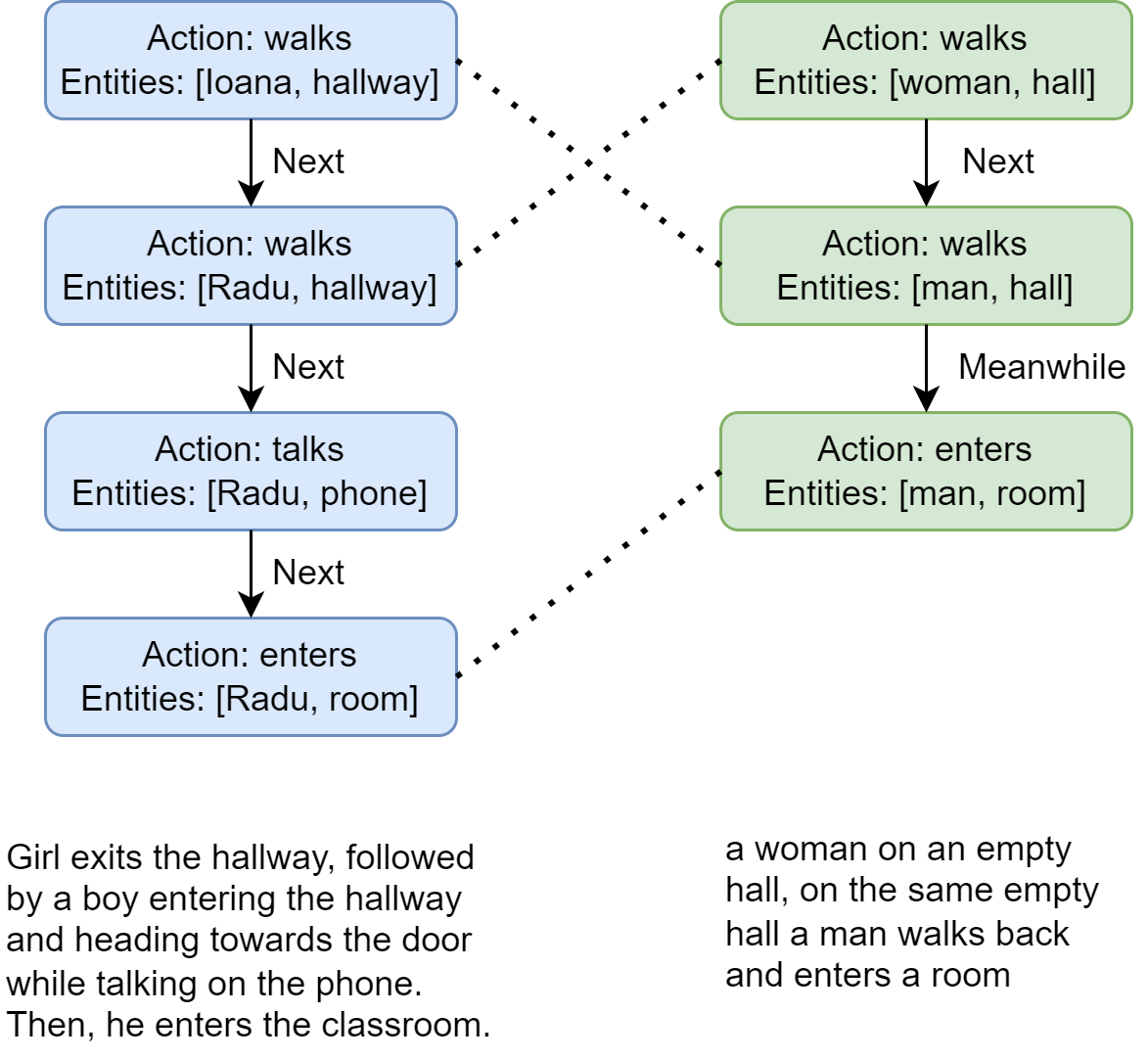}
    \caption{Pair of GEST and text that stem from the same video. Dotted lines mark matched nodes. GEST graph matching score: 0.3493, prediction: 1. BLEURT similarity: 0.3755, prediction: 0.}
    \label{fig:diff_graph_bleurt_1}
\end{figure}

\begin{figure}
    \centering
    \includegraphics[width=\columnwidth]{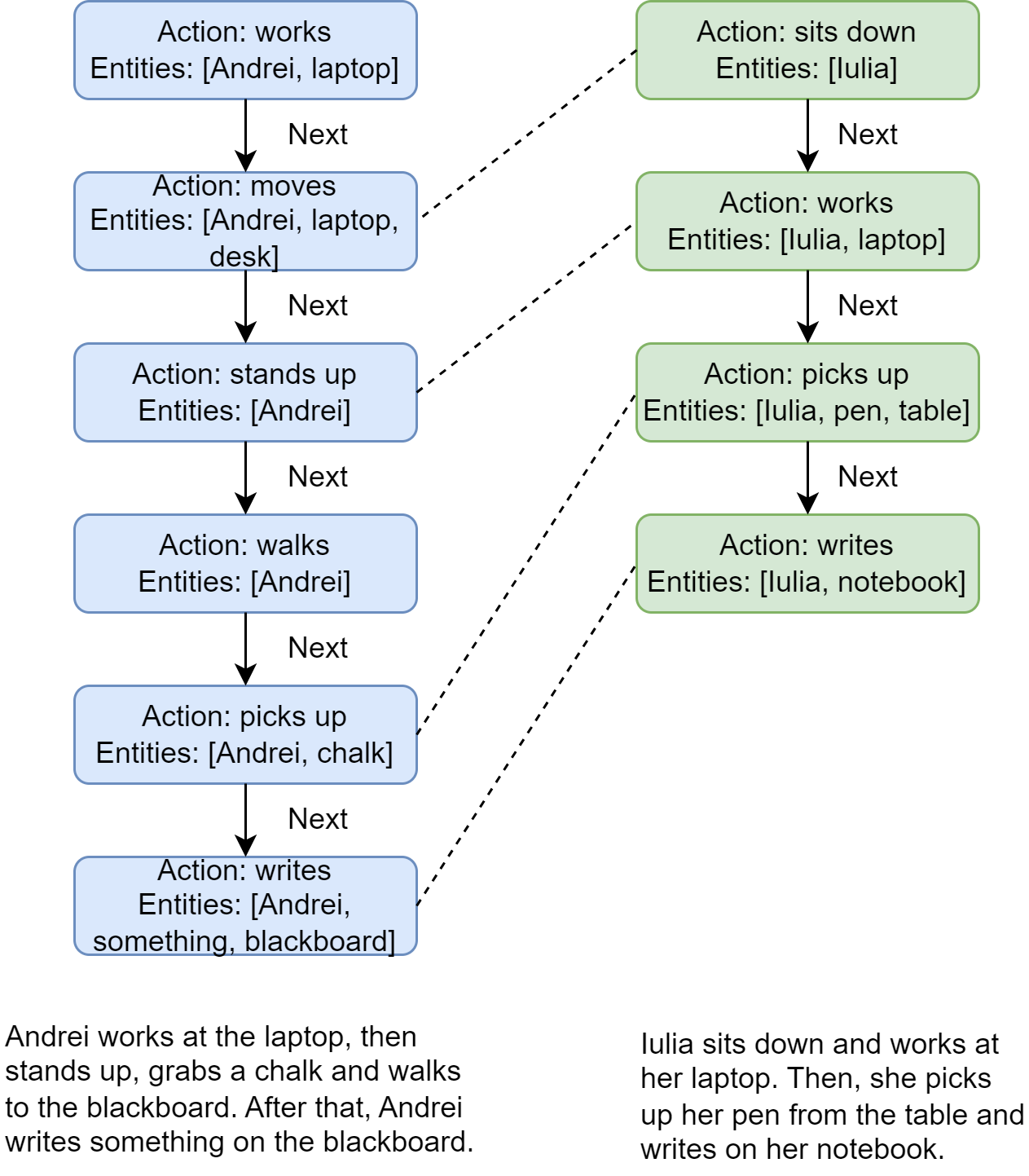}
    \caption{Pair of GEST and text that stem from the different videos. Dotted lines mark matched nodes. GEST graph matching score: 0.0822, prediction: 0. BLEURT similarity: 0.5625, prediction: 1.}
    \label{fig:diff_graph_bleurt_0}
\end{figure}

\section{GEST serialization}

\begin{figure*}
    \centering
    \includegraphics[width=\textwidth]{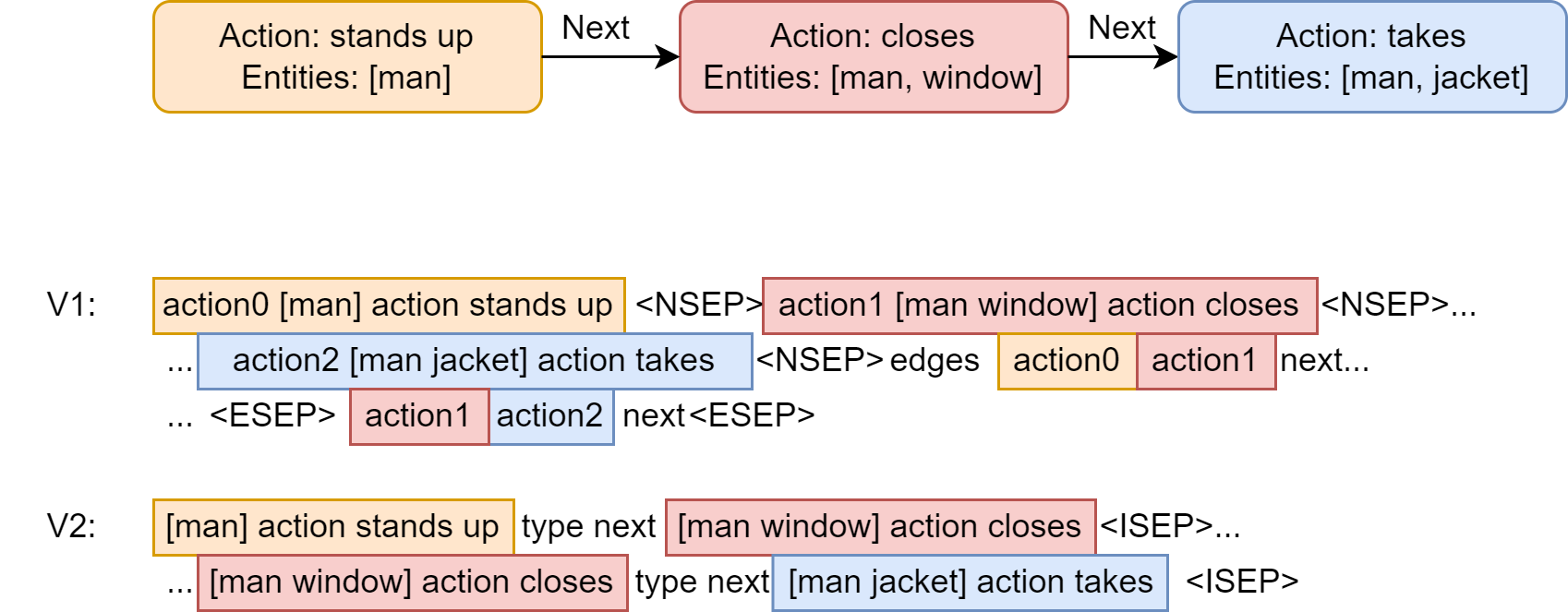}
    \caption{Methodologies for transforming a GEST into a string.}
    \label{fig:stringified_graph}
\end{figure*}

In order to be used either as input or output in a training pipeline, the graph representation needs to be transformed into the format used by the employed encoder-decoder models. For our case, this is represented by natural language text. Besides the need to be as close as possible to natural language, the serialized graph should contain all the information from the original graph and ensure that the original is recoverable from the serialized version. Following~\cite{ribeiro2020investigating} and after extensive experimenting with different serialization methods, we settle for two methodologies, one for each task (i.e. graph-to-text and text-to-graph). For text-to-graph task we opt to use a process that generates string that are easier to fix (so they are sound) and simpler to generate. In the case of graph-to-text generation, we prefer a richer representation that reduces the number of references, thus easing the learning process (searching for a particular reference in text while easy to solve programmatically, is a hard task for Transformer based encoder-decoder models. Both methodologies (V1 is used for text-to-graph while V2 for is used for graph-to-text). are depicted on a illustrative example in Figure~\ref{fig:stringified_graph}.

\section{GEST Graph Mathcing vs BLEURT}
\label{section:graph_vs_bleurt}

In Figures~\ref{fig:diff_graph_bleurt_1} and~\ref{fig:diff_graph_bleurt_0} we present two examples of pairs from the Videos-to-Paragraphs dataset. In Figure~\ref{fig:diff_graph_bleurt_1}, both entries (graph + text) stem from the same video, information that is reproduced by the graph matching approach. BLEURT metric fails to capture this information due, most probably, to the different writing style of the two texts. While the first one is descriptive, the second one is richer, more complex (e.g. "same empty hall") and significantly shorter. For the example in Figure~\ref{fig:diff_graph_bleurt_1}, BLEURT incorrectly marks the texts as stemming from the same video. This highlights a limitation of the metric, as it is fails to understand that while the actions might look similar, they are still very different, especially when considering they are performed by different actors. Our approach is not focused exclusively on actions, as it also take into account the entities involved in a certain action. The action writing with a pen in a notebook while semantically similar, is still different from writing on a blackboard. Breaking up the action and entities involved allows for a finer semantic level of detail, as we will be comparing actions with actions and entities with entities. 

Note that we take a very simple approach to building the affinity matrix, that coupled with graph matching algorithms does not always yield optimal results. Furthermore, the metric (affinity matrix and graph matching algorithm) is not trained or optimized for this task or dataset. Even with this very basic approach and limited data, we obtain state-of-the-art results for text similarity, proving the power of GEST. We leave optimizing (e.g. by error analysis) the affinity matrix and learning the metric for further research.

\end{document}